\pdfoutput=1

\documentclass[11pt]{article}

\usepackage[final]{acl}

\usepackage{times}
\usepackage{latexsym}
\usepackage{amsmath}

\usepackage[T1]{fontenc}

\usepackage[utf8]{inputenc}

\usepackage{microtype}

\usepackage{inconsolata}

\usepackage{graphicx}

\title{Predicting Customer Satisfaction by Replicating the Survey Response Distribution}

\author{Etienne Manderscheid, Matthias Lee\\
        Dialpad Canada Inc. \\
        1100 Melville St \#400 \\ 
        Vancouver, BC, Canada, V6E 4A6 \\
  \texttt{\{etienne,matthias.lee\}@dialpad.com}}

\begin{document}
\maketitle
\begin{abstract}
For many call centers, customer satisfaction (CSAT) is a key performance indicator (KPI). However, only a fraction of customers take the CSAT survey after the call, leading to a biased and inaccurate average CSAT value, and missed opportunities for coaching, follow-up, and rectification. Therefore, call centers can benefit from a model predicting customer satisfaction on calls where the customer did not complete the survey. Given that CSAT is a closely monitored KPI, it is critical to minimize any bias in the average predicted CSAT (pCSAT). In this paper, we introduce a method such that predicted CSAT (pCSAT) scores accurately replicate the distribution of survey CSAT responses for every call center with sufficient data in a live production environment. The method can be applied to many multiclass classification problems to improve the class balance and minimize its changes upon model updates.
\end{abstract}

\section{Introduction}
Many machine learning applications use classifiers updated periodically by developers. Without special control mechanisms, these updates can shift the relative balance of output classes, causing unintended effects. For the case of predicting CSAT, we developed a control mechanism to address this issue, taking care to mitigate the risks posed by sampling noise. This paper explains our method and strategies for handling sampling noise, and aims to help developers seeking to replicate one or more target class distribution(s). 

Customer satisfaction (CSAT) is critical for call center performance assessment, yet often measured through surveys completed by a small subset of customers—averaging 8\% in our dataset. This limited response rate can skew perceived performance, as non-responding customers' satisfaction remains unknown. Predicting CSAT for all calls, even those without survey responses, can mitigate this bias \citet{mand_lee_2023}.

Ensuring these pCSAT scores do not introduce further bias is crucial. This paper introduces a method to more accurately replicate the distribution of survey CSAT responses in a live production environment, addressing limitations identified in prior work and providing more accurate metrics for call center performance.

\section{Related Work}
Predicting CSAT using machine learning models has gained attention, especially in call center conversations. The challenge is not only predicting accurate scores but also ensuring these predictions replicate the true distribution of survey responses. This section reviews relevant studies and methodologies applied to similar problems, focusing on distribution replication and ordinal classification (since CSAT is measured on a 1-5 scale).

\subsection{Predicting Customer Satisfaction}
Previous research explored various approaches to predict CSAT from contact center conversations. \citet{Bockhorst2017} developed a system using ASR-generated call transcripts, non-textual data, and sentiment scores to predict a Representative Satisfaction Index (RSI) with rank scoring and isotonic regression models. Similarly, \citet{auguste2019} used the Net Promoter Score (NPS) with binary classification (promoters vs. detractors)  for predicting customer satisfaction in chat conversations, achieving moderate improvements with a macro F1 score of 53.8\%.

Other studies examined predicting CSAT from raw audio signal features such as acoustic, emotional, and prosodic features \citep{park2009towards, zweig2006automated, vaudable2012negative, devillers2010real}.

This work builds on a previously developed method for predicting CSAT scores using ASR-generated (Automated Speech Recognition) call transcripts \citep{mand_lee_2023}. This method improved prediction accuracy with a mapping function from binary model outputs to 5 CSAT classes (Figure \ref{fig:mappings1}). The mapping function was parameterized by 4 decision thresholds. The binary model itself was a trained Big bird, a transformer with sparse attention optimized to handle long input sequences (such as call transcripts) with a linear memory requirement \citep{zaheer2020bigbird}.

\subsection{Replicating Class Distribution}
Our threshold fitting approach replicates the survey CSAT distribution, crucial for maintaining class proportions in predictions. Research on maintaining class distribution intersects with imbalanced learning and ordinal regression, using techniques like resampling, re-weighting, and threshold adjustment to handle class imbalances.
Model calibration can be a helpful addition to these methods, but is not a replacement, as model calibration focuses on adjusting predicted probabilities to better reflect true likelihoods, which does not imply that the class distribution will be faithfully replicated if the decision thresholds are incorrect.

\textbf{Re-sampling and Re-weighting}: These techniques adjust training processes to account for class imbalances, ensuring minority classes are represented. However, they are not well suited to replicating an exact class balance, as the effects of these training set adjustments are difficult to predict. 

\textbf{Threshold Optimization in Multiclass and Ordinal Classification}: Threshold optimization is critical in contexts requiring precise class predictions, such as multiclass and ordinal classification tasks. \citet{kotsiantis2006} discuss methods to adjust decision thresholds for imbalanced datasets, balancing sensitivity and specificity to represent minority classes. 
\citealp{ferri2002} introduce methods to optimize decision thresholds to minimize misclassification costs. Their work is relevant in contexts where different misclassifications have different costs, making threshold adjustment crucial. While these approaches are closely related to ours by adjusting model thresholds to reflect true class distributions, they focus on binary and multiclass classification without emphasizing ordinal classes as ours does.

\citet{cardoso2007} proposed a data replication method for ordinal classification, handling ordinal data by replicating instances to indirectly optimize thresholds for ordinal predictions. This study aligns with our work, emphasizing maintaining the natural order of classes, but our method directly optimizes thresholds to replicate survey responses, rather than using data replication.

In "A simple approach to ordinal classification," \citealp{frank2001} propose a threshold-based method for ordinal classification problems. Their approach involves training a series of binary classifiers to predict whether an instance belongs to a class above a certain threshold. This method is closely related to our approach, as both aim to predict ordinal classes by optimizing thresholds. However, we use a single classifier, and our method goes further by ensuring that the predicted class distribution matches the training class distribution, a step beyond the basic ordinal classification task.

\citet{chu2007} explored ordinal regression using support vector machines (SVMs), optimizing thresholds within the SVM framework to respect the ordinal nature of data. Similar to our work, this study focuses on ordinal data and threshold optimization, but our method is model-agnostic, post-processing outputs of any classifier to match desired distributions.

These studies provide valuable insights into threshold optimization for multiclass and ordinal classification. Our work distinguishes itself by:
\begin{enumerate}
    \item Optimizing specific decision thresholds to align pCSAT scores with CSAT survey responses, ensuring calibration and class distribution replication. 
    \item Creating a custom loss function reflecting our unique product goals and user suggestions. 
    \item Applying our method to a large language model (LLM) predicting CSAT scores from call center conversations, integrating threshold optimization into a broader machine learning pipeline to address practical challenges in real-world settings.
\end{enumerate}

\section{Data \& Methods}
\subsubsection{Transcripts}
We used conversational transcripts generated from our Automatic Speech Recognition engine. The accuracy (1 - Word Error Rate) was > 85\%.

\subsubsection{Calls}
We used approximately 892K call center calls with a CSAT survey score and a model-assigned pCSAT probability ranging from June 24, 2023 to June 17, 2024.

\subsubsection{Trials}
To rule out effects due to chance or periodicity, we ran the experiment 7 times using different training and test periods. The last of those trials corresponds to a production deployment of the model, and the other trials were simulated for the purposes of this analysis. Each trial consists of a 60 day training period and 120 day test period. A 30 day period separates the start of one trial from the start of the next (thus trials overlap). We expected and observed no differences between the deployed and simulated trials since the pipeline is the same. 

\subsubsection{Call Centers}
We excluded call centers with fewer than 5 high and 5 low CSAT calls over the 60-day training period to avoid very high sampling noise. To better understand the impact of sampling noise, we further categorized call centers heuristically based on the number of survey CSAT responses in the 60-day training period. Table \ref{table:call_centers_distribution} shows the number of call centers in each survey response bin, summed over the 7 trials. 

\begin{table}
\begin{tabular}{p{3cm} p{3cm}}
\hline
\hline
\textbf{Survey Responses} & \textbf{\# of Call Centers} \\ \hline
1-50                               & 401                             \\ \hline
51-200                             & 908                             \\ \hline
201-500                            & 425                             \\ \hline
501-1000                           & 199                             \\ \hline
> 1000                             & 197                             \\ \hline
\end{tabular}
\caption{Number of Call Centers by Survey Responses Volume}
\label{table:call_centers_distribution}
\end{table}

\subsection{Threshold Optimization Procedure}
\subsubsection{Model and Mapping}
\begin{figure}[h]
  \centering
  \includegraphics[width=1\linewidth]{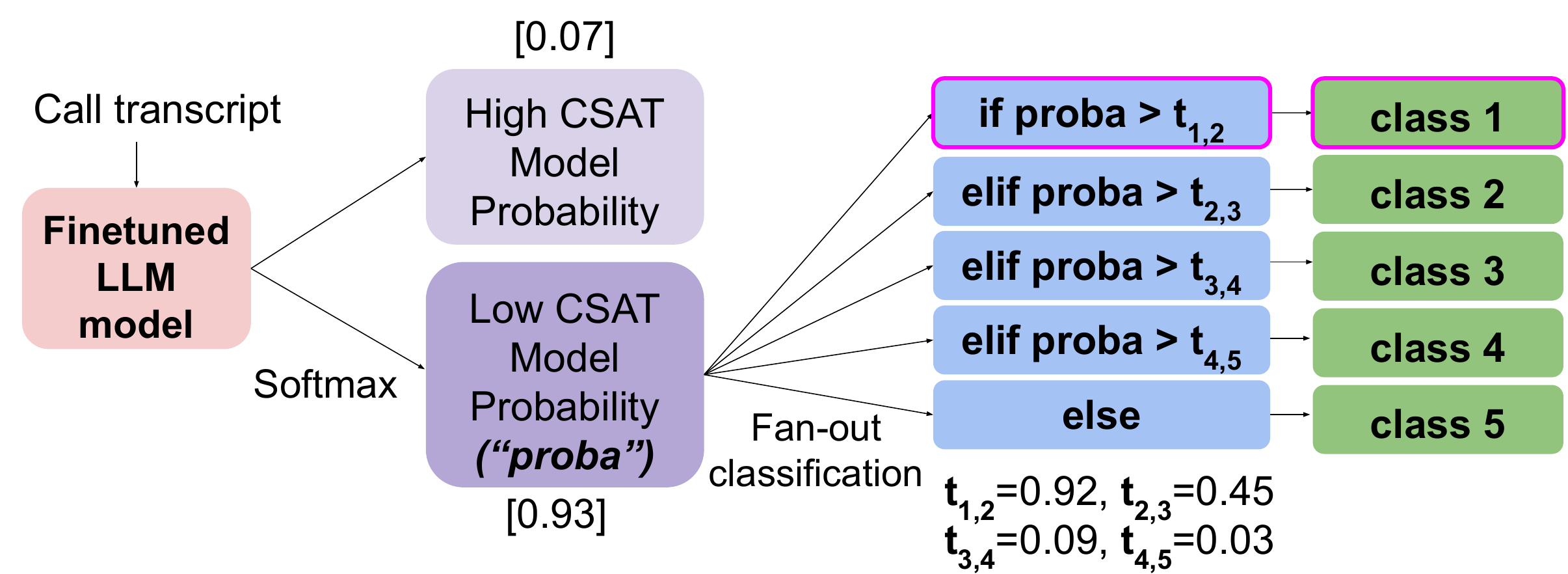}
  \caption{The mapping function that takes \textit{low CSAT} probability (\textbf{"proba"}) as input and outputs 1-5 pCSAT. In this example, the \textbf{"proba"} is 0.93, which is larger than \( t_{1,2} \) so the model emits a pCSAT of 1.}
  \label{fig:mappings1}
\end{figure}

The model is a large language model (LLM) that predicts CSAT with binary outputs: high or low CSAT. Details on the model and how it was trained are provided in \citealp{mand_lee_2023}. The model also provides the probability of both classes, referred to as \textbf{"proba"} for the \textit{low CSAT} class. The mapping function (Figure \ref{fig:mappings1}) uses this probability to output a pCSAT score on a 1-5 scale. The mapping has four parameters representing decision thresholds (i.e. class boundaries): \( t_{1,2} \), \( t_{2,3} \), \( t_{3,4} \), and \( t_{4,5} \). For example, \( t_{3,4} \) is the probability threshold separating a pCSAT of 3 from 4.

\subsubsection{Product Requirements}
Our approach is based on meeting product requirements, ranked by importance:
\begin{enumerate}
    \item The average pCSAT should equal the average survey CSAT over the same set of calls. By default, the displayed average pCSAT (and average CSAT) is  the \% of satisfied calls, i.e., calls with pCSAT $\geq$ 4. 
    \item These averages should also match when toggled to use a 1-5 scale. 
    \item The distribution of pCSAT and survey CSAT should match as closely as possible
\end{enumerate}
Based on customer feedback we set 1\% and 0.1 as the maximum differences to target for requirements 1 and 2 respectively.

\subsubsection{Parameter Estimation}

\paragraph{Jointly estimate the four thresholds:}
To find the optimal parameters, our process iterates through different combinations of thresholds to find the set that minimizes the loss. 

To meet all three product requirements, we created the following loss function:

\[
\textrm{Loss} = \Delta\%_{p, c} + \Delta \textrm{avg}_{p, c} + \textrm{MSE}_{p, c}
\]
where \textit{p} and \textit{c} are short for pCSAT and CSAT, respectively, and:

\[
\Delta\%_{p, c} = |\left( \% \textrm{ of pCSAT} \geq 4 \right) - \left( \% \textrm{ of CSAT} \geq 4 \right)|
\]

\[
\Delta \textrm{avg}_{p, c} = |\mathrm{avg\_pcsat} - \mathrm{avg\_csat}|
\]

\[
\textrm{MSE}_{p, c} = \textrm{MSE}(\vec{pcsat}, \vec{csat})
\]

We preferred a random search to a grid search to save computation time. We used 5000 iterations for random search, but found 98.7\% convergence by 500 iterations. 

\subsubsection{Optimization Steps:}
\begin{enumerate}
    \item Compute the Number of Calls for Each Survey CSAT Level:
    \[
    \mathrm{n}_{\mathrm{csat_i}} = \textrm{number of calls with survey CSAT}
    \]
    where class $i \in (1,2,3,4,5)$

    \item Calculate the Average Survey CSAT:
    \[
    \mathrm{avg\_csat} = \frac{\sum_{i=1}^{5} (\mathrm{n}_{\mathrm{csat_i}} \cdot i)}{\sum_{i=1}^{5} \mathrm{n}_{\mathrm{csat_i}}}
    \]

    \item Initialize the loss:
    \[
    \mathrm{best\_loss} = 1000.0
    \]

    \item Random Search for Optimal Thresholds:
    Perform a random search through the possible thresholds to find the set that minimizes the loss.
    For $j$ in range(5000):
    \begin{enumerate}
        \item Generate 4 uniform random values and sort them:
        \[
\mathrm{t_{12} > t_{23} > t_{34} > t_{45}} \sim \mathcal{U}(0, 1)
\]

    \item Compute the Number of Calls for Each pCSAT Level:
    \[
    \mathrm{n}_{\mathrm{pcsat_i}} = \textrm{number of calls with pCSAT}
    \]
    where class $i \in (1,2,3,4,5)$

    \item Calculate the Average pCSAT:
    \[
    \mathrm{avg\_pcsat} = \frac{\sum_{i=1}^{5} (\mathrm{n}_{\mathrm{pcsat_i}} \cdot i)}{\sum_{i=1}^{5} \mathrm{n}_{\mathrm{pcsat_i}}}
    \]

    \item Compute the Delta Between Average pCSAT and CSAT:
    \[
    \Delta\mathrm{pcsat\_csat} = \mathrm{avg\_pcsat} - \mathrm{avg\_csat}
    \]

    \item Compute \(\Delta\%_{p, c}\)
    
    \item Normalize both CSAT Vectors to unit length:
    \[
    \vec{pcsat} = \text{normalized}([\mathrm{n}_{\mathrm{pcsat_1}}, \ldots ,\mathrm{n}_{\mathrm{pcsat_5}}])
    \]
    \[
    \vec{csat} = \text{normalized}([\mathrm{n}_{\mathrm{csat_1}}, \ldots ,\mathrm{n}_{\mathrm{csat_5}}])
    \]

    \item Calculate the Mean Squared Error Between the Normalized Vectors
    (\( \mathrm{MSE}_{p,c} \))
    \item Compute the loss:
    
    \[
    \textrm{Loss} = \Delta\%_{p, c} + \Delta \textrm{avg}_{p, c} + \textrm{MSE}_{p, c}
    \]

    \item Update the loss and best thresholds if the current loss is lower.
\end{enumerate}
\end{enumerate}

\subsection{Experimental Conditions}
We evaluated the loss under five conditions:
\begin{enumerate}
    \item \textbf{\textcolor[rgb]{0.2588,0.5216,0.9569}{Baseline}:} Naive model output (evaluated over test period)
    \item \textbf{\textcolor[rgb]{0.929,0.263,0.208}{Global Threshold}:} Thresholds are fitted on a single, global pool (evaluated over test period)
    \item \textbf{\textcolor[rgb]{0.984,0.737,0.016}{Call Center Threshold}:} Individual thresholding for each call center  (evaluated over test period)
    \item \textbf{\textcolor[rgb]{0.204,0.659,0.325}{Train Period}:} We apply the same call center-specific parameters as used in the \textcolor[rgb]{0.984,0.737,0.016}{Call Center Threshold} condition, but apply them to the training period instead of the test period. As we expect a near-zero difference of means, this serves to validate our parameter estimation method.
    \item \textbf{\textcolor[rgb]{1.000,0.427,0.004}{Bootstrap (Train Period)}:} This approach is similar to the "train" condition, but the key difference is that we repeatedly resample the training set and measure the difference of means over these samples. This method helps us estimate how much of the loss in the "Call center thresholding" condition is due to sampling noise, and attribute the rest to differences between train and test distributions, i.e. model drift.
\end{enumerate}
We note that the first 3 conditions are test conditions, i.e. evaluated on the test set, whereas the last 2 are train conditions which help us understand sources of error. 

\section{Results}
The effectiveness of different methods for predicting customer satisfaction (CSAT) scores was evaluated through various experimental conditions. The results are summarized in Figures 2-5, and detailed observations are as follows:

\subsection{Loss}
\begin{figure}[h]
\centering
\includegraphics[width=1\linewidth]{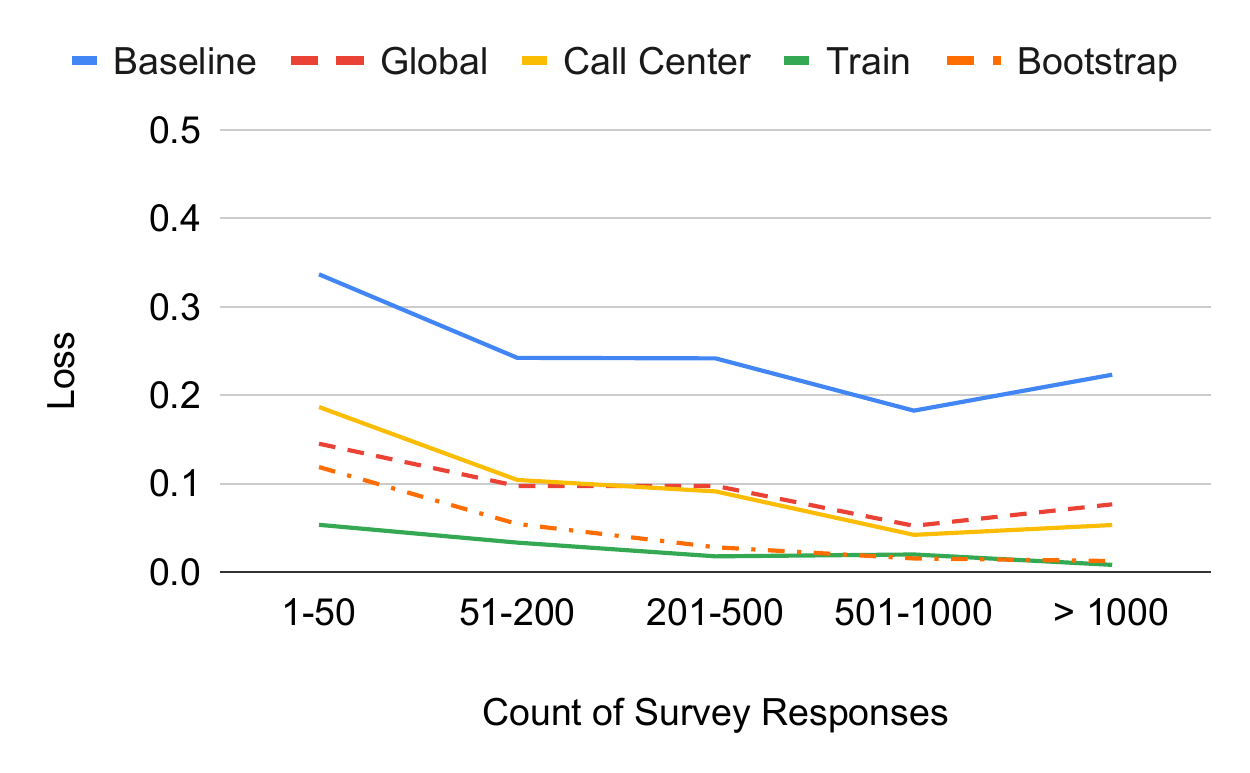}
\caption{The average loss for each of the five experimental conditions, binned by the call center's CSAT survey responses.}
\label{fig:figure2}
\end{figure}

Overall loss, depicted in Figure \ref{fig:figure2}, combines the difference of means, difference of percent satisfied, and MSE. We see that the \textcolor[rgb]{0.2588,0.5216,0.9569}{Baseline} method consistently has the highest loss, and the \textcolor[rgb]{0.204,0.659,0.325}{Train} condition has the lowest. The train condition does not have 0 error because it is not usually possible to find 4 parameters to zero all 3 terms that make up the loss simultaneously. 

Now we compare the test conditions. For call centers with the smallest response volumes, the \textcolor[rgb]{0.929,0.263,0.208}{Global Threshold} method performs best. On the other hand, the \textcolor[rgb]{0.984,0.737,0.016}{Call Center Threshold} method performed best for call centers with the largest response volumes. Overall, we see a gradual trend of this method improving as the response volume increases. This makes sense since the \textcolor[rgb]{0.984,0.737,0.016}{Call Center Threshold} method is limited by sampling noise, which is greatest for small response volumes. Indeed, we can see the effect of the sampling noise directly by looking at how much more loss the \textcolor[rgb]{1.000,0.427,0.004}{Bootstrap} condition has relative to the \textcolor[rgb]{0.204,0.659,0.325}{Train} condition at low  (< 200) response volumes. As we get to higher (> 500) response volumes, we observed that the \textcolor[rgb]{1.000,0.427,0.004}{Bootstrap} condition catches up with the \textcolor[rgb]{0.204,0.659,0.325}{Train} condition, which indicates that sampling error ceases to be significant at those volumes. 

\subsection{Difference in \% of Satisfied Calls}
\begin{figure}[h]
\centering
\includegraphics[width=1\linewidth]{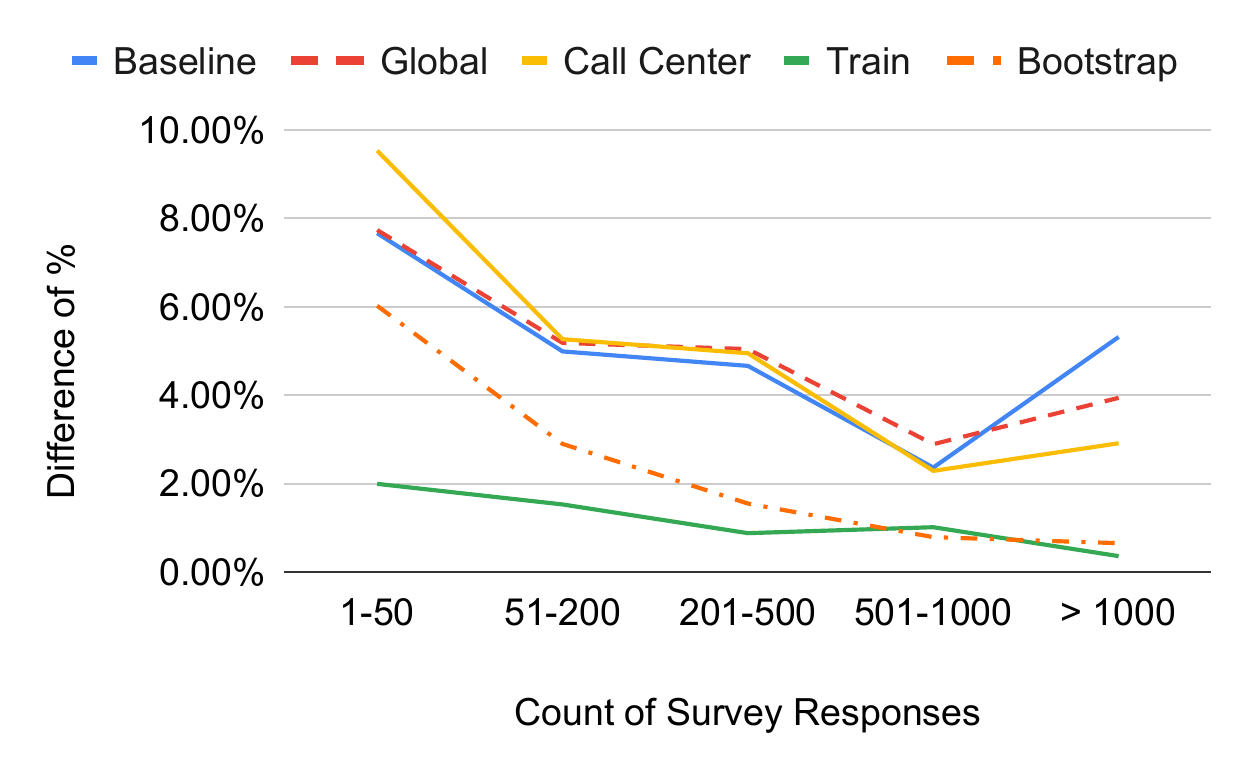}
\caption{The average difference in percentage of satisfied calls between pCSAT and CSAT, broken down by the call center’s count of survey responses.}
\label{fig:figure3_100324}
\end{figure}

Figure \ref{fig:figure3_100324} examines the first component of the loss function, the difference in percentage of satisfied calls between pCSAT and CSAT, averaged over the call centers in each bin. The \textcolor[rgb]{0.2588,0.5216,0.9569}{Baseline} method was only outperformed for call centers with the largest response volume (> 1000). For these call centers, the \textcolor[rgb]{0.984,0.737,0.016}{Call Center Threshold} method performs best, followed by \textcolor[rgb]{0.929,0.263,0.208}{Global Threshold}. We also note an unexpected uptick in error for all 3 test conditions, compared to smaller response volumes. 

\subsection{Difference of Means}
\begin{figure}[h]
\centering
\includegraphics[width=1\linewidth]{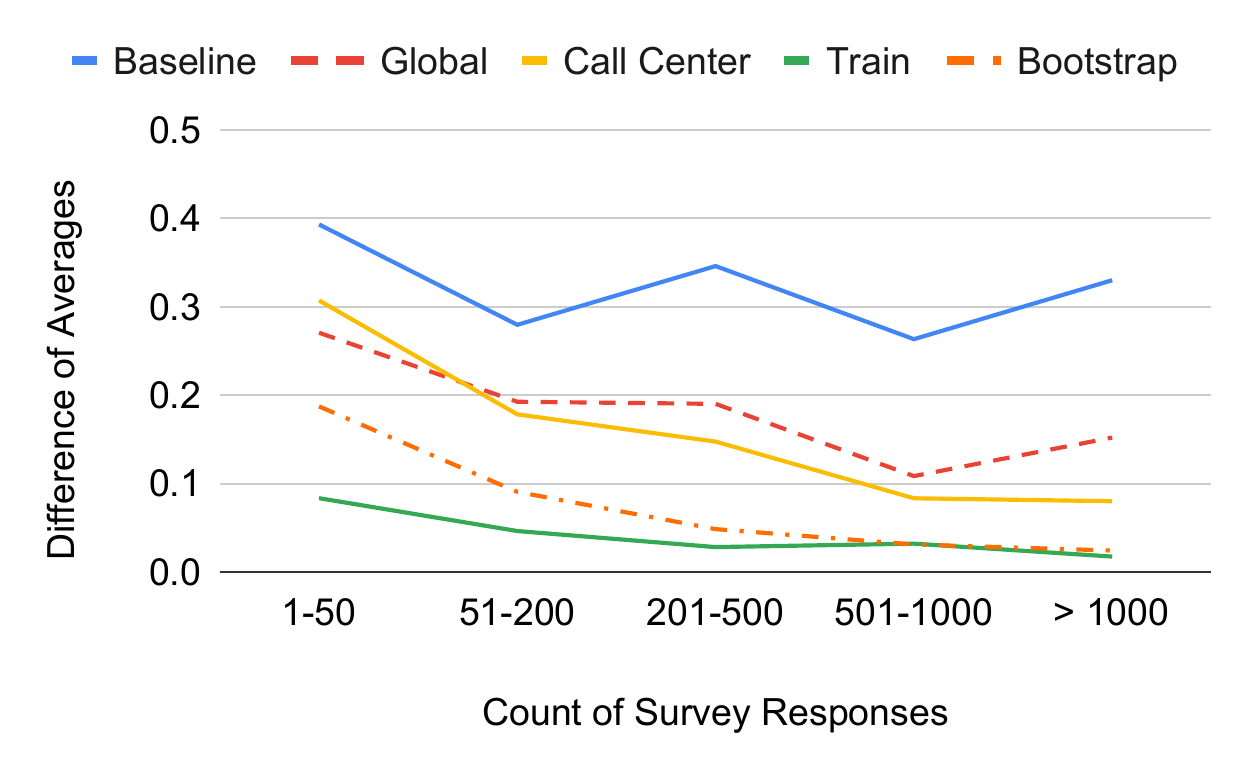}
\caption{The average absolute difference between mean Predicted CSAT and survey CSAT for each of the five experimental conditions, binned by the call center's CSAT survey responses.}
\label{fig:figure4}
\end{figure}

Figure \ref{fig:figure4} focuses on the average absolute difference between mean Predicted CSAT and survey CSAT. The \textcolor[rgb]{0.2588,0.5216,0.9569}{Baseline} method consistently lags other methods, showing our methods create a substantial improvement. 
The \textcolor[rgb]{0.929,0.263,0.208}{Global Threshold} method performs best at the lowest response volumes (< 200), whereas the \textcolor[rgb]{0.984,0.737,0.016}{Call Center Threshold}  method outperforms other methods from 200 calls onwards, consistent with its requirement of small sampling noise. 

As expected, the \textcolor[rgb]{0.204,0.659,0.325}{Train} and \textcolor[rgb]{1.000,0.427,0.004}{Bootstrap} conditions show very low percentages, further validating the parameter estimation and highlighting the minimal impact of sampling noise after 500 calls.

\subsection{MSE}
\begin{figure}[h]
\centering
\includegraphics[width=1\linewidth]{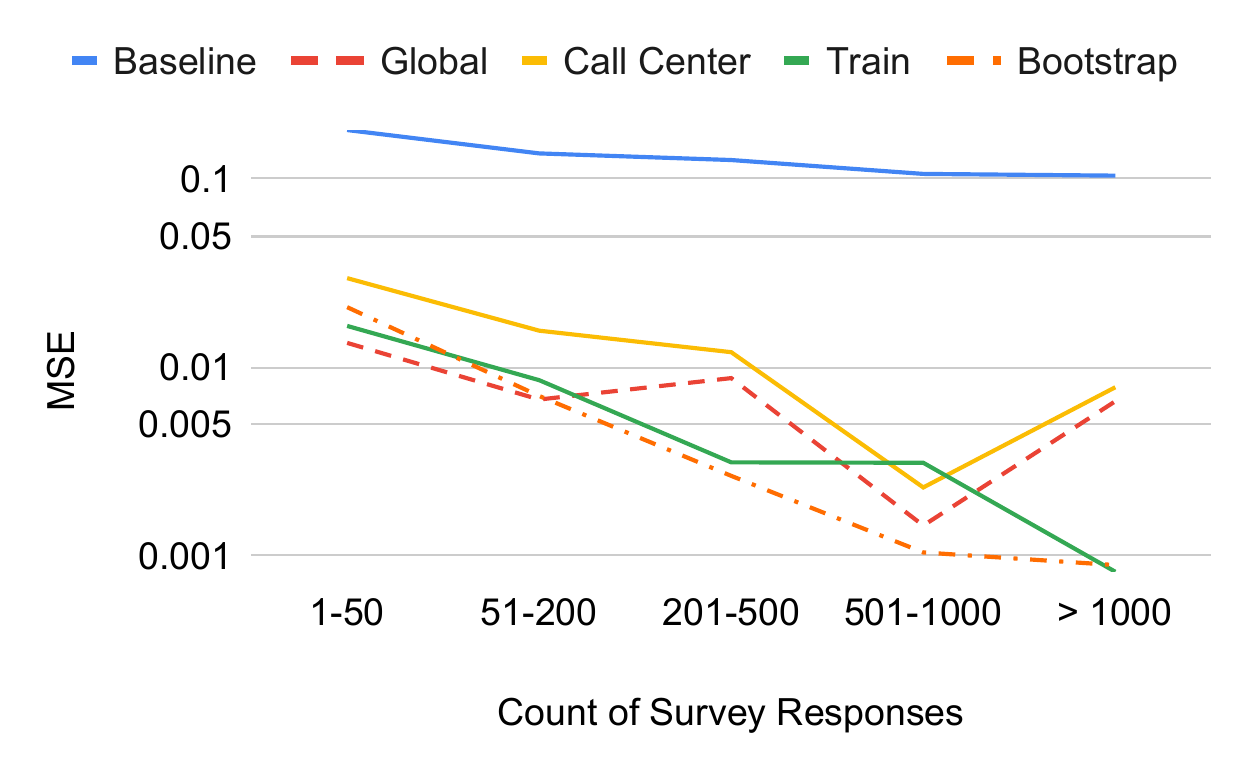}
\caption{The Mean Squared Error measures the vector alignment between the normalized Predicted CSAT and survey CSAT distributions. Shown for each of the five experimental conditions, binned by the call center's CSAT survey responses.}
\label{fig:figure5}
\end{figure}

The MSE, shown in \ref{fig:figure5}, measures the vector similarity between the pCSAT and CSAT distributions. 
The \textcolor[rgb]{0.2588,0.5216,0.9569}{Baseline} method exhibits by far the highest MSE values across all call volumes, so much so the figure requires a logarithmic scale. It exceeds 0.1 for all survey response volumes. 
The \textcolor[rgb]{0.929,0.263,0.208}{Global Threshold} method is consistently best at lowering the MSE, though the gap with \textcolor[rgb]{0.984,0.737,0.016}{Call Center Threshold} narrows as survey responses increase.

\section{Discussion}
Some but not all initial targets were achieved:
\begin{itemize}
    \item \textbf{Difference in Percentage of Satisfied Calls:} Not achieved. The methods were not able to improve over \textcolor[rgb]{0.2588,0.5216,0.9569}{Baseline} or meet the target of less than 1\%. The fact the fitting thresholds does not improve the output class distribution suggests that the baseline classifier outputs may be already well distributed. This is a likely explanation because this metric treats CSAT as binary (call satisfied if CSAT >= 4) and the baseline classifier is trained on binary CSAT data. Further work may be required to yield improvements.
    \item \textbf{Difference of Means:}  The \textcolor[rgb]{0.984,0.737,0.016}{Call Center Threshold} method consistently achieved the target (difference of means less than 0.1) for call centers with survey responses greater than 500-1000. This method also achieved significant improvements over \textcolor[rgb]{0.2588,0.5216,0.9569}{Baseline} for all call center bins. 
    \item \textbf{Mean Squared Error (MSE):} Both the \textcolor[rgb]{0.929,0.263,0.208}{Global Threshold} and \textcolor[rgb]{0.984,0.737,0.016}{Call Center Threshold}  methods significantly improved the MSE over \textcolor[rgb]{0.2588,0.5216,0.9569}{Baseline}, indicating improved alignment between pCSAT and actual CSAT distributions. In this case, there was no quantitative target, but the improvement is over 10X. 
\end{itemize}

Having learned from the varying performance of the methods across different survey response volumes, we are now considering implementing a hybrid approach. Specifically, using the \textcolor[rgb]{0.929,0.263,0.208}{Global Threshold} method for call centers with fewer than 200 survey responses and \textcolor[rgb]{0.984,0.737,0.016}{Call Center Threshold} method for larger ones. This hybrid strategy leverages the strengths of both methods, ensuring more accurate and reliable pCSAT predictions across diverse operational contexts.

We recommend a similar approach for multiclass classification problems where a consistent class balance is important across model updates. Our approach can be used whether there is a single pool of inputs, or subgroups analogous to our call centers. Developers should be cautious of the sampling noise in their datasets and use a data-driven approach to determine the  minimum sample sizes for their specific application.

\section{Limitations}
While our thresholding method demonstrates substantial improvements, several limitations must be acknowledged:
\begin{itemize}
    \item \textbf{Sampling Noise:} As highlighted, small call centers with low survey response volumes suffer from high sampling noise, limiting the effectiveness of our approach for sample sizes under ~500, especially call center thresholding.
    \item \textbf{Temporal Stability:} Although our method shows promise in maintaining low loss over at least 4 months, we did not examine the timecourse of errors over those 4 months or beyond. Long-term drift could be a concern and warrants further investigation.
\end{itemize}

\section{Ethics Statement}
In developing and implementing this method, we have adhered to ethical standards to ensure fairness, transparency, and accountability:
\begin{itemize}
    \item \textbf{Bias Mitigation:} Previously, we have sampled subpopulations of users and evaluated internally to ensure the pCSAT is not biased against specific groups. This approach takes a further step to reduce bias in pCSAT scores by ensuring a more accurate reflection of customer satisfaction across different call centers. However, continuous evaluation and improvement are necessary to address any emergent biases, and our near-term plans include quantifiable and verifiable explainability for AI CSAT which will help our users pinpoint the causes of low pCSAT, including any bias.
    \item \textbf{Transparency:} We have documented our methods and findings comprehensively to provide clear insights into our process and its impact on prediction accuracy.
    \item \textbf{Data Privacy:} All customer data used in this study has been anonymized and handled in compliance with data privacy regulations to protect individual privacy. We follow the data privacy measures in place at Dialpad which include scrubbing personal identifiable information (PII) from customer data and restricting our use of customer data to improvements to the services we provide them. We did not rely on any external annotations.
    \item \textbf{Stakeholder Impact:} The improved accuracy in CSAT predictions enables better decision-making for coaching, follow-up, and service improvements, ultimately benefiting customers and call center performance.
\end{itemize}

\section{Acknowledgements}
We would like to acknowledge the contributions of our colleagues, the support from Dialpad, and the feedback from call center managers who helped refine our approach. In particular, we thank person Doug Mackenzie for maintaining essential database tables.


\bibliography{custom}  

\begin{thebibliography}{13}
\providecommand{\natexlab}[1]{#1}

\bibitem[{Auguste et~al.(2019)Auguste, Charlet, Damnati, Bechet, and Favre}]{auguste2019}
Jeremy Auguste, Delphine Charlet, Geraldine Damnati, Frederic Bechet, and Benoit Favre. 2019.
\newblock Can we predict self-reported customer satisfaction from interactions?
\newblock In \emph{Proceedings of the 2019 Conference on Neural Information Processing Systems (NeurIPS)}, pages 7385--7389.

\bibitem[{Bockhorst et~al.(2017)Bockhorst, Yu, Polania, and Fung}]{Bockhorst2017}
Joseph Bockhorst, Shi Yu, Luisa Polania, and Glenn Fung. 2017.
\newblock Predicting self-reported customer satisfaction of interactions with a corporate call center.
\newblock In \emph{Machine Learning and Knowledge Discovery in Databases}, pages 179--190, Cham. Springer International Publishing.

\bibitem[{Cardoso and da~Costa(2007)}]{cardoso2007}
J.~S. Cardoso and J.~F.~Pinto da~Costa. 2007.
\newblock Learning to classify ordinal data: The data replication method.
\newblock \emph{Journal of Machine Learning Research}, 8:1393--1429.
\newblock Retrieved from jmlr.org (SpringerLink).

\bibitem[{Chu and Keerthi(2007)}]{chu2007}
W.~Chu and S.~S. Keerthi. 2007.
\newblock \href {https://doi.org/10.1162/neco.2007.19.3.792} {Support vector ordinal regression}.
\newblock \emph{Neural Computation}, 19(3):792--815.
\newblock Retrieved from doi.org (SpringerLink).

\bibitem[{Devillers et~al.(2010)Devillers, Vaudable, and Chastagnol}]{devillers2010real}
Laurence Devillers, Caroline Vaudable, and Chantal Chastagnol. 2010.
\newblock Real-life emotion-related states detection in call centers: a cross-corpora study.
\newblock In \emph{Eleventh Annual Conference of the International Speech Communication Association (INTERSPEECH)}, volume~10, pages 2350--2353.

\bibitem[{Ferri et~al.(2002)Ferri, Hernández-Orallo, and Salido}]{ferri2002}
C.~Ferri, J.~Hernández-Orallo, and M.~A. Salido. 2002.
\newblock Learning decision trees using the area under the roc curve.
\newblock In \emph{Proceedings of the 19th International Conference on Machine Learning}, pages 139--146.
\newblock Retrieved from researchgate.net.

\bibitem[{Frank and Hall(2001)}]{frank2001}
E.~Frank and M.~Hall. 2001.
\newblock A simple approach to ordinal classification.
\newblock In \emph{European Conference on Machine Learning}, pages 145--156.
\newblock Retrieved from springer.com.

\bibitem[{Kotsiantis et~al.(2006)Kotsiantis, Kanellopoulos, and Pintelas}]{kotsiantis2006}
S.~B. Kotsiantis, D.~Kanellopoulos, and P.~E. Pintelas. 2006.
\newblock Handling imbalanced datasets: A review.
\newblock \emph{GESTS International Transactions on Computer Science and Engineering}, 30(1):25--36.
\newblock Retrieved from gests-intl.com.

\bibitem[{Manderscheid and Lee(2023)}]{mand_lee_2023}
E.~Manderscheid and M.~Lee. 2023.
\newblock Predicting customer satisfaction with soft labels for ordinal classification.
\newblock In \emph{Proceedings of the 61st Annual Meeting of the Association for Computational Linguistics (Volume 5: Industry Track)}, pages 652--659, Toronto, Canada. Association for Computational Linguistics.
\newblock Anonymized for review.

\bibitem[{Park and Gates(2009)}]{park2009towards}
Young-Bum Park and Stephen~C. Gates. 2009.
\newblock Towards real-time measurement of customer satisfaction using automatically generated call transcripts.
\newblock In \emph{Proceedings of the 18th ACM Conference on Information and Knowledge Management}, pages 1387--1396. ACM.

\bibitem[{Vaudable and Devillers(2012)}]{vaudable2012negative}
Caroline Vaudable and Laurence Devillers. 2012.
\newblock Negative emotions detection as an indicator of dialogs quality in call centers.
\newblock In \emph{2012 IEEE International Conference on Acoustics, Speech and Signal Processing (ICASSP)}, pages 5109--5112. IEEE.

\bibitem[{Zaheer et~al.(2020)Zaheer, Guruganesh, Dubey, Ainslie, Alberti, Ontanon, Pham, Ravula, Wang, Yang, and Ahmed}]{zaheer2020bigbird}
Manzil Zaheer, Guru Guruganesh, Avinava Dubey, Joshua Ainslie, Chris Alberti, Santiago Ontanon, Philip Pham, Anirudh Ravula, Qifan Wang, Li~Yang, and Amr Ahmed. 2020.
\newblock Big bird: Transformers for longer sequences.
\newblock \emph{arXiv preprint arXiv:2007.14062}.
\newblock Submitted on 28 Jul 2020 (v1), last revised 8 Jan 2021 (v2).

\bibitem[{Zweig et~al.(2006)Zweig, Siohan, Saon, Ramabhadran, Povey, Mangu, and Kingsbury}]{zweig2006automated}
Geoffrey Zweig, Olivier Siohan, George Saon, Bhuvana Ramabhadran, Daniel Povey, Lidia Mangu, and Brian Kingsbury. 2006.
\newblock Automated quality monitoring for call centers using speech and nlp technologies.
\newblock In \emph{Proceedings of the 2006 Conference of the North American Chapter of the Association for Computational Linguistics on Human Language Technology: Companion Volume: Demonstrations}, pages 292--295. Association for Computational Linguistics.

\end{thebibliography}

\end{document}